\documentclass[conference,a4paper]{IEEEtran}
\IEEEoverridecommandlockouts

\usepackage{cite}
\usepackage{amsmath,amssymb,amsfonts}
\usepackage{algorithm}
\usepackage{algpseudocode}
\usepackage{graphicx}
\usepackage{booktabs}
\usepackage{multirow}
\usepackage{url}
\usepackage[hidelinks]{hyperref}

\graphicspath{{./}}

\newcommand{\ARS}{\mathrm{ARS}}
\newcommand{\code}[1]{\texttt{#1}}

\newcommand{\authorentry}[2]{%
  \parbox[t]{0.30\textwidth}{\centering
    {\normalsize\bfseries #1}\\[-1pt]
    {\small #2}}}
\begin{document}

\title{Trust Responsive Runtime Authorization for Agentic AI with Empirical Characterization of Policy Behavior and Parameter Sensitivity}

\author{%
\begin{tabular*}{\textwidth}{@{\extracolsep{\fill}}ccc}
\authorentry{Ahsan Habib Shihab*}{Farabi University, Almaty, Kazakhstan\\ahsan.h.shihab@gmail.com\\
ORCID: 0009-0008-1157-6335} &
\authorentry{Stefan Gubo}{Faculty of Economics and Informatics\\
Selye Janos University, Komarno, Slovakia.\\} &

\authorentry{Ladiszlav Vegh}{Faculty of Economics and Informatics\\
Selye Janos University, Komarno, Slovakia.\\
} \\
\noalign{\vskip 12pt}

\authorentry{Imre Felde}{Obuda University, Budapest, Hungary.\\
Faculty of Economics and Informatics\\
Selye Janos University, Komarno, Slovakia.\\
} &

\authorentry{Berik Sabdenaliyev}{SDU University, Almaty, Kazakhstan.\\
Obuda University, Budapest, Hungary.} &

\authorentry{Zhasulan Akylaev}{Farabi University, Almaty, Kazakhstan.} \\
\noalign{\vskip 12pt}

\authorentry{Ramis Akhmedov}{SDU University, Almaty, Kazakhstan.\\
Doctoral School of Applied Informatics and Applied Mathematics,\\ Obuda University, Budapest, Hungary.} &

\authorentry{Madina Mansurova}{Farabi University, Almaty, Kazakhstan.} &

\authorentry{Amir Mosavi}{Obuda University, Budapest, Hungary.\\
Ludovika University of Public Service, Budapest, Hungary.\\
Selye Janos University, Slovakia.\\
Farabi University, Almaty, Kazakhstan\\
Abylkas Saginov Karaganda Technical University, Karaganda, Kazakhstan.\\
Astana IT University, Kazakhstan.\\
Qazaq AI Research University, Kazakhstan.\\
}
\end{tabular*}}

\IEEEaftertitletext{\vspace{-1.5\baselineskip}}
\maketitle

\begin{abstract}
The policy that converts runtime estimates of an autonomous agent's condition
into operational authority is rarely specified or measured. We empirically
characterize the Trust-Responsive
Authorization Policy (TRAP), which maps an Authorization Readiness Score (ARS), a
weighted combination of runtime trust, operational stability and contextual risk,
onto discrete authorization levels through an asymmetric rule: escalation
requires consecutive observations above a band edge plus a margin, while a
sharp drop in readiness restricts authority at once. On one matched set of 100
Monte Carlo traces, we compare TRAP with static, trust-only, risk-only and
symmetric policies. Under abrupt degradation, TRAP restricts at onset and
re-escalates sooner after recovery than the symmetric policy, with the highest
and, descriptively, least variable stability among the adaptive policies,
although more of its post-recovery escalations occur below the selected trust
threshold. Paired bootstrap intervals exclude zero for the remaining differences. In one-factor
sweeps, recovery time
rises with the escalation margin and run length, shifts with the score weights,
and is unchanged by the restriction threshold. In an exploratory post-freeze
experiment, when the same degradation is spread over 5--20 intervals, TRAP restricts only on the 4 traces that retain a
single-interval drop above that threshold; averaged over all 100 traces, it holds
authority above what readiness supports for 15.9--18.9 of the 20 degradation
intervals,
against 0.8 under abrupt degradation and about 10 for the symmetric policy. Its
restriction advantage therefore depends on degradation being abrupt. All
reported values are traceable to the publicly available, versioned research
release.
\end{abstract}


\section{Introduction}

Agentic AI systems often operate in environments whose conditions change during
execution. \emph{Runtime assessment} (estimating trust, monitoring behavior,
quantifying risk) establishes what an agent's current condition is [1].
Authorization is a different decision: how much operational authority the agent
should hold in that condition. We study the policy that makes this second
decision [2]. The Trust-Responsive Authorization Policy (TRAP) introduces \emph{Authorization
Readiness} as an intermediate decision abstraction, instantiates it as an
Authorization Readiness Score (ARS) which combines runtime trust, operational
stability and contextual risk, and converts that score into a discrete
authorization level through an \emph{asymmetric} rule: authority is accumulated
conservatively and withdrawn immediately. Implementing this
principle requires the policy to carry state between decisions. The work is an empirical characterization of a stateful,
graded-authority transition policy on identical runtime traces; it does not
propose another general access-control or enforcement framework [3]. This study makes four contributions:
\begin{enumerate}
\item An implementation-level specification of TRAP (explicit initialization,
ordered transition branches, counter semantics and restriction target),
sufficient to reproduce the evaluated policy (Section~\ref{sec:method},
Algorithm~\ref{alg:trap}); its readiness score, band structure and
asymmetric-threshold rule are inherited from~\cite{companion}.
\item A matched-trace comparison of TRAP with four alternative policies under
abrupt degradation, characterizing restriction and recovery latency,
authorization stability and post-recovery escalation behavior (exp010,
Section~\ref{sec:res-behavior}).
\item A one-factor sensitivity analysis across $13$ sweep points, showing that
recovery time increases with the escalation margin and required run length,
remains unchanged across the tested restriction-threshold values and varies
non-monotonically with the score weights (exp011,
Section~\ref{sec:res-sensitivity}).
\item An additional exploratory, post-freeze analysis of gradual degradation, showing that
TRAP's immediate-restriction advantage depends on the rate of readiness decline
and largely disappears under the tested gradual ramps, leaving authority above
the readiness-mapped level for longer than under the symmetric policy (exp012,
Section~\ref{sec:res-gradual}).
\end{enumerate}
The scope is one policy on one synthetic testbed; we characterize how TRAP
behaves and do not offer a general methodology for designing runtime
authorization policies.

\section{Related Work}
\label{sec:related}

Access control has moved from static roles~\cite{rbac} and
attributes~\cite{abac} to decisions re-evaluated during access~\cite{ucon},
risk-based~\cite{radac} or verified per request~\cite{zerotrust}. These decide
whether an access is permitted; TRAP, complementing them, decides how much
authority an executing agent holds.
Trust in automation should track observed reliability~\cite{leesee,hoff}, with
adaptive~\cite{okamura}, longitudinal~\cite{devisser} or
uncertainty-aware~\cite{dutta} calibration; authorization policies consume such
estimates. Agentic and LLM-based systems are delegated growing
scope~\cite{agenticsurvey,llmagents}; TRAP's levels follow levels of
automation~\cite{parasuraman} and autonomy levels for AI
agents~\cite{autonomylevels}; runtime enforcement~\cite{agentspec},
delegation~\cite{delegation} and governance
frameworks~\cite{nistrmf,oecd,trustworthyai} call for continuous oversight.
None of these works specifies, or characterizes the behavior of, a policy that
moves an executing agent between graded authority levels as its condition
changes. Our earlier study~\cite{companion} defined TRAP's asymmetric-threshold rule
and introduced PDI, a structural measure of how authority at a fixed condition
depends on history; it measured no restriction, recovery or stability behavior,
varied no policy parameters and tested no degradation, which this paper does.

\section{Policy Definition}
\label{sec:method}

The policy is stated as implemented in \code{policies.py} and \code{sim.py}.
TRAP is the asymmetric-threshold rule analysed structurally
in~\cite{companion}; Algorithm~\ref{alg:trap} states it in complete operational
form.
Authorization Readiness, the degree to which an agent is currently prepared
to hold authority, is instantiated for trust $T_t$, stability $S_t$ and risk
$R_t$ at decision interval $t$ as
\begin{equation}
g_t \;=\; \ARS(T_t,S_t,R_t) \;=\; w_0 T_t + w_1 S_t - w_2 R_t ,
\label{eq:ars}
\end{equation}
All results use $w=(0.5,\,0.3,\,0.2)$. The linear form keeps each component's
contribution explicit; the policy layer, not the readiness estimator, is under
study. Authority takes six ordered levels $a \in \{0,\dots,5\}$ induced by
\begin{equation}
\mathcal{B} = (0.20,\; 0.35,\; 0.50,\; 0.65,\; 0.80),
\label{eq:bands}
\end{equation}
via $\sigma(g) = |\{b \in \mathcal{B} : g \ge b\}|$, with upper edge
$\mathrm{up}(a) = \mathcal{B}_a$ for $a<5$ ($+\infty$ otherwise).
The levels are ordinal and abstract: level $0$ is the least and level $5$ the
most operational authority, and the simulator attaches no concrete permissions
to them. Assigning permissions to levels, for example which actions an agent
may take without approval, is a deployment decision in the sense of autonomy
levels~\cite{parasuraman,autonomylevels} and is not evaluated here. Table~\ref{tab:notation} summarizes the notation and the
baseline settings. The score weights, the band edges and $\theta_e$, $\theta_r$
and $k$ are design values carried over unchanged from~\cite{companion}, not
empirically calibrated or optimized; Section~\ref{sec:res-sensitivity} examines
how the results depend on the weights and thresholds.

\begin{table}[t]
\caption{Notation and baseline settings. Every reported result uses these
values unless a sensitivity sweep varies one of them.}
\label{tab:notation}
\centering
\scriptsize
\setlength{\tabcolsep}{3pt}
\begin{tabular}{@{}lll@{}}
\toprule
\textbf{Symbol} & \textbf{Meaning} & \textbf{Baseline} \\
\midrule
$T_t, S_t, R_t$ & trust, stability, risk at interval $t$ & in $[0,1]$ \\
$g_t$ & Authorization Readiness Score, Eq.~\eqref{eq:ars} & --- \\
$w=(w_0,w_1,w_2)$ & score weights & $(0.5,0.3,0.2)$ \\
$\mathcal{B}$, $\sigma(g)$ & band edges; level mapped from $g$ & Eq.~\eqref{eq:bands} \\
$a_t$ & authorization level & $\{0,\dots,5\}$ \\
$\mathrm{up}(a)$, $\mathrm{low}(a)$ & upper, lower edge of level $a$ & --- \\
$c_t$ & consecutive escalation count & $c_0=0$ \\
$\theta_e$ & escalation margin above $\mathrm{up}(a)$ & $0.05$ \\
$\theta_r$ & restriction threshold on a one-step drop & $0.10$ \\
$k$ & required consecutive observations & $3$ \\
\midrule
$H$ & intervals per trace & $100$ \\
$t_F$, $t_R$ & degradation, recovery onset & $40$, $60$ \\
$n$, seed & traces per run; generator seed & $100$, $42$ \\
$L$ & degradation ramp length (exp012) & $1,5,10,20$ \\
\midrule
AS, RR, RT & stability; restriction, recovery delay & Sec.~\ref{sec:design} \\
PER, PRR & escalation shares (degradation; $T_t<0.60$) & Sec.~\ref{sec:design} \\
EAD & intervals with $a_t>\sigma(g_t)$ in $[t_F,t_R)$ & Sec.~\ref{sec:design} \\
\bottomrule
\end{tabular}
\end{table}

TRAP maintains $(a_t,c_t)$: the level and a counter of consecutive qualifying
escalation observations, initialized as $a_0=\sigma(g_0)$, $c_0=0$.
Algorithm~\ref{alg:trap} gives the transition. Each interval is processed in
three ordered cases. \emph{Restriction}: if readiness falls by more than $\theta_r$ from the
previous interval, the level becomes
$\min(a_{t-1},\sigma(g_t))$ and the counter resets. \emph{Escalation}:
otherwise, if $g_t\ge\mathrm{up}(a_{t-1})+\theta_e$, the counter increments; the
level remains unchanged until the counter reaches $k$, when it rises by one and
the counter resets. \emph{Hold}: otherwise, the level remains unchanged and the
counter resets. Thus, restriction responds to a single large drop, while
escalation requires $k$ consecutive qualifying observations. One consequence follows directly from the
rule: the level can fall only through the restriction branch, so a decline in
which no single-interval drop exceeds $\theta_r$ never lowers it, however far
readiness falls. Section~\ref{sec:res-gradual} measures this consequence.

\begin{algorithm}[t]
\caption{TRAP transition (per decision interval)}
\label{alg:trap}
\begin{algorithmic}[1]
\Require $T_t,S_t,R_t$; previous state $(a_{t-1},c_{t-1})$; previous score
$g_{t-1}$; parameters $w,\theta_e,\theta_r,k$
\Ensure updated state $(a_t,c_t)$
\State $g_t \gets w_0T_t + w_1S_t - w_2R_t$
\State $d_t \gets g_t - g_{t-1}$
\If{$d_t < -\theta_r$} \Comment{immediate restriction}
  \State $a_t \gets \min(a_{t-1}, \sigma(g_t))$; \; $c_t \gets 0$
\ElsIf{$g_t \ge \mathrm{up}(a_{t-1}) + \theta_e$} \Comment{conservative escalation}
  \State $c_t \gets c_{t-1} + 1$
  \If{$c_t \ge k$}
    \State $a_t \gets a_{t-1} + 1$; \; $c_t \gets 0$
  \Else
    \State $a_t \gets a_{t-1}$
  \EndIf
\Else \Comment{hold}
  \State $a_t \gets a_{t-1}$; \; $c_t \gets 0$
\EndIf
\State \Return $(a_t, c_t)$
\end{algorithmic}
\end{algorithm}

\noindent\textbf{Comparison policies.}
Four alternatives consume the same traces and level map. Static fixes
$a_t=\sigma(g_0)$. Symmetric uses the same $k$-consecutive mechanism in
both directions (at or above $\mathrm{up}(a_{t-1})+\theta$ to rise, below
$\mathrm{low}(a_{t-1})-\theta$ to fall, with $\mathrm{low}(a)=\mathcal{B}_{a-1}$),
which isolates the effect of asymmetry. Separate rise and fall counters are
kept; meeting one condition resets the other counter, and meeting neither resets
both. At level $0$ the implementation uses a lower edge of $-10^{9}$, so the
level cannot fall and the fall counter stays at $0$; as implemented it is parameterized by
$\theta_e$ and does not consume $\theta_r$. Trust-only and
risk-only apply the TRAP rule to the single signal $T_t$ or $1-R_t$,
isolating the effect of composing three signals.

\section{Supporting Structural Background}
\label{sec:pd}

We summarize from the structural analysis reported in~\cite{companion} only what
Section~\ref{sec:res-pdi} needs. Because TRAP carries $(a,c)$, its output
depends on history, not on $(T_t,S_t,R_t)$ alone. For a terminal condition
$x^\star$, let $\mathcal{H}(x^\star)$ be the set of all admissible histories
ending exactly at $x^\star$,
$\widehat{\mathcal{H}}(x^\star)\subseteq\mathcal{H}(x^\star)$ the finite subset
produced by the generator below, and $\lambda(h)$ the level TRAP assigns at the
end of $h$. The Path-Dependence Index (PDI) and the history-sensitive fraction
(HSF) are
\begin{align}
\mathrm{PDI}(x^\star) &= \max_{h \in \mathcal{H}(x^\star)} \lambda(h)
                       - \min_{h \in \mathcal{H}(x^\star)} \lambda(h),
\label{eq:pdi}\\[2pt]
\mathrm{HSF} &= \frac{\left|\{x^\star \in \mathcal{X} : \mathrm{PDI}(x^\star) > 0\}\right|}
                     {\left|\mathcal{X}\right|},
\label{eq:hsf}
\end{align}
where $\mathcal{X}$ is the finite set of target conditions, the measured
$\widehat{\mathrm{PDI}}(x^\star)$ is \eqref{eq:pdi} evaluated over
$\widehat{\mathcal{H}}(x^\star)$, and
$\widehat{\mathrm{HSF}}=|\{x^\star\in\mathcal{X}:\widehat{\mathrm{PDI}}(x^\star)>0\}|/|\mathcal{X}|$.
PDI counts the bands separating the most
and least authorized of observationally identical agents; it is not an
expectation over natural histories. Restricting \eqref{eq:pdi} to a subset can
only narrow the spread, so
$\widehat{\mathrm{PDI}}(x^\star)\le\mathrm{PDI}(x^\star)$ and
$\widehat{\mathrm{HSF}}\le\mathrm{HSF}$: every measured spread is a \emph{lower
bound} on the full-history PDI.

\label{sec:envelope}\label{sec:reach}\label{sec:generator}
\emph{Admissible} means no faster than the $99.99$th percentile of the
simulator's own ordinary per-step changes, measured in exp004 as
\begin{equation}
\Delta T_{\max} = 0.0632, \qquad \Delta S_{\max} = 0.0421 .
\label{eq:env}
\end{equation}
with $R$ unbounded per step; given the observed value ranges ($T,S \le 1$,
$R \ge 0.150$), the score cannot exceed $g=0.770$. The generator builds each history of $\widehat{\mathcal{H}}(x^\star)$ in three phases (escalate to a
requested carried level, hold it, descend to $x^\star$) and checks with
executable assertions that the history ends exactly at $x^\star$, respects the
envelope, carries the requested level, and regenerates bit-identically. During
escalation the generator holds $S=0.75$ and $R$ at its minimum, an admissible but
not maximal choice that matters in Section~\ref{sec:res-pdi}.

\section{Experimental Design}
\label{sec:design}

\paragraph{Runtime model}
The simulator generates $H=100$-interval traces in three scripted phases:
stable operation, induced degradation at $t_F=40$, and recovery onset at
$t_R=60$. Trust and stability
start from $\mathcal{U}(0.60,0.75)$ and $\mathcal{U}(0.65,0.85)$ and take Gaussian
steps with standard deviation $0.015$ and $0.010$, drop once at $t_F$ by
$\mathcal{U}(0.20,0.40)$ and $\mathcal{U}(0.10,0.25)$, and from $t_R$ gain an extra
$\mathcal{U}(0.005,0.015)$ and $\mathcal{U}(0.003,0.010)$ per interval; risk is
drawn from $\mathcal{U}(0.15,0.30)$, $\mathcal{U}(0.60,0.85)$ and
$\mathcal{U}(0.30,0.50)$ in the three phases. Per interval, the Gaussian step
($t \ge 1$), drop ($t=t_F$) and drift ($t \ge t_R$) are applied in that order,
then risk is drawn and trust and stability are clipped to $[0,1]$.

\paragraph{Traces and the seed}
We call one generated input sequence $(T_t,S_t,R_t)_{t<H}$ a \emph{trace}; a
policy's level sequence $(a_t)$ on a trace is its response, and every metric is
computed per trace. A run draws $n=100$ traces sequentially from one NumPy
\code{default\_rng} stream seeded with $42$, so the $100$ traces are distinct,
independent draws, not repetitions of one trace. Trace generation does not
depend on the policy, and each run restarts the stream from seed $42$. Every
policy, and every sensitivity setting, is therefore evaluated on the same $100$
traces. This matched design is deliberate: differences between policies or
settings cannot arise from their inputs, and they can be compared trace by
trace.

\paragraph{Gradual degradation (exp012)}
The frozen experiments contain only the abrupt drop at $t_F$. exp012 is an
exploratory, post-freeze experiment; its recorded protocol states that it was
written after the exp010 and exp011 results and the property of
Section~\ref{sec:method} were known, but before exp012 was run. The experiment
keeps every random draw of each trace and changes only the
schedule on which the drawn degradation is applied. The trust and stability
drops are subtracted in $L$ equal parts at $t_F,\dots,t_F+L-1$, and over the
same intervals the bounds of the risk distribution move linearly from the
stable to the degraded range, with the same uniform draw. Risk must be ramped as
well, because its step at $t_F$ alone lowers $g$ by about $0.1$, close to
$\theta_r$. We use $L\in\{5,10,20\}$; $L=20$ ends at $t=59$, and the frozen
generator is the abrupt control ($L=1$). A trace meets the intended gradual
condition if no one-interval decline of $g$ inside the ramp exceeds $\theta_r$.
The protocol also defines the metric EAD below and records predictions, which
are reported against the outcome.

\paragraph{Behavioral metrics}
\label{sec:metrics}
With $\mathcal{I}=\{t\ge1 : a_t>a_{t-1}\}$ the escalation intervals, five metrics
are computed per trace.
\begin{itemize}
\item \textbf{Authorization Stability (AS)}
$= 1 - |\{t \ge 1 : a_t \ne a_{t-1}\}| / (H-1)$, the fraction of intervals at
which authority did not change. It measures stability, not safety: a level held
while readiness falls also counts as stable, so AS is read together with RR
and, in exp012, EAD.
\item \textbf{Restriction Responsiveness (RR)}
$= \min\{t \ge t_F : a_t < a_{t-1}\} - t_F$; $\mathrm{RR}=0$ means restriction in
the degradation interval itself.
\item \textbf{Recovery Time (RT)} $= \min\{t \ge t_R : a_t > a_{t-1}\} - t_R$.
\item \textbf{Protective Efficiency Ratio (PER)}
$= |\{t \in \mathcal{I} : t_F \le t < t_R\}| / |\mathcal{I}|$, the share of
escalations granted during degradation (lower is better).
\item \textbf{Progressive Recovery Ratio (PRR)}
$= |\{t \in \mathcal{I}, t \ge t_R : T_t < 0.60\}| /
|\{t \in \mathcal{I} : t \ge t_R\}|$, the share of post-recovery escalations
granted while trust is below $0.60$ (lower is better). The value $0.60$ is a
fixed evaluation threshold selected for this study, not an operationally validated
trust boundary.
\item \textbf{Excess-authority duration (EAD)}, in exp012
$= |\{t_F \le t < t_R : a_t > \sigma(g_t)\}|$, the number of the $20$
degradation intervals in which the level exceeds the one readiness maps to
(lower is better). The reference $g_t$ uses the baseline weights for every
policy.
\end{itemize}
Event metrics are averaged over the traces in which the event occurs; a policy
that never changes level has them undefined and reported as n/a, not zero.
EAD is defined on every trace.

\paragraph{Experiments}
Table~\ref{tab:experiments} lists the experiments in the shared repository. The
primary experiments of this paper are exp010, the policy comparison
(Table~\ref{tab:baseline}, Fig.~\ref{fig:trajectory}), and exp011, the
sensitivity sweeps (Table~\ref{tab:sensitivity}, Fig.~\ref{fig:sensitivity}).
exp012 (Table~\ref{tab:gradual}) was added after the freeze, outside the frozen
tree.

\begin{table}[t]
\caption{Experiments, each with runner and stored results. exp003--exp009 come
from the earlier structural study and supply supporting evidence; exp003--exp011
are in the frozen repository; exp012 was added after the freeze.}
\label{tab:experiments}
\centering
\scriptsize
\setlength{\tabcolsep}{3pt}
\begin{tabular}{@{}llr@{}}
\toprule
\textbf{ID} & \textbf{Role} & \textbf{Scale} \\
\midrule
exp003 & Multi-level condition prevalence     & $5\times5$ settings, 4000 ep.\ each \\
exp004 & Admissibility envelope               & 20{,}000 episodes \\
exp005 & Generator verification (pilot)       & 264 histories \\
exp006 & Pilot path-dependence (null result)  & 50 targets \\
exp007 & Reachable memory enumeration         & BFS, 500 score bins \\
exp008 & Generator verification (final)       & 200 histories \\
exp009 & Primary path-dependence measurement  & 50 targets $\times$ 2 seeds \\
\midrule
exp010 & Policy behavior comparison (primary) & 5 policies $\times$ 100 traces \\
exp011 & Parameter sensitivity (primary)      & 13 sweep points   \\
\midrule
exp012 & Gradual degradation (post-freeze)  & 4 profiles $\times$ 5 policies \\
\bottomrule
\end{tabular}
\end{table}

\paragraph{Prespecified criteria}
\label{sec:prereg}
The condition grid and falsification criteria for the path-dependence
measurement belong to~\cite{companion} and were fixed in a frozen protocol
before any value was computed: $50$ conditions selected by occurrence from the prevalence mining of
Section~\ref{sec:res-prevalence} (bin width $0.02$), with no property of the
emitted level used for selection. H1 was to be rejected if $\widehat{\mathrm{HSF}}<0.10$
and mean $\widehat{\mathrm{PDI}}<0.10$; H2 if mean $\widehat{\mathrm{PDI}}$ were non-monotone in
$k\in\{2,3,4\}$. H2 was never tested: no PDI sweep over $k$ exists, and
our $k$ sweep measures behavioral metrics, not PDI.

\section{Results}
\label{sec:results}

Every point value is read from stored artifacts. The values of
Sections~\ref{sec:res-behavior}--\ref{sec:res-sensitivity} were additionally
re-derived from the policy implementation, and the principal statistics of
Section~\ref{sec:res-pdi} are re-derived by the reproduction harness. Interval
estimates are 95\% percentile bootstrap intervals ($10{,}000$ resamples over
traces, one fixed seed) from per-run records verified to reproduce every stored
aggregate (Section~\ref{sec:repro}); differences are paired trace by trace, over
the traces on which both values are defined. They describe sampling variation of
the trace generator only and are not adjusted for multiple comparisons.

\subsection{Policy Behavior}
\label{sec:res-behavior}

\begin{table}[t]
\caption{Policy behavior on identical runtime traces ($n=100$, seed $42$,
$w=(0.5,0.3,0.2)$, $\theta_e=0.05$, $\theta_r=0.10$, $k=3$). AS is reported as
mean\,$\pm$\,SD over all $100$ traces. Event metrics average over the traces on
which they are defined: RR over $99/100/98/100$ and RT, PER, PRR over
$95/100/99/100$ traces for trust-only/risk-only/symmetric/TRAP. ``n/a'' denotes
a metric undefined on every trace. For PER and PRR, lower is better.}
\label{tab:baseline}
\centering
\footnotesize
\begin{tabular}{@{}lccccc@{}}
\toprule
\textbf{Policy} & \textbf{AS} & \textbf{RR} & \textbf{RT} & \textbf{PER} & \textbf{PRR} \\
\midrule
Static      & $1.0000 \pm 0.0000$ & n/a  & n/a    & n/a    & n/a    \\
Trust-only  & $0.9673 \pm 0.0100$ & 0.00 & 15.13  & 0.0096 & 0.5053 \\
Risk-only   & $0.9385 \pm 0.0148$ & 0.00 & \phantom{0}2.28 & 0.2405 & 0.7950 \\
Symmetric   & $0.9646 \pm 0.0104$ & 2.00 & 17.14  & 0.0000 & 0.4798 \\
TRAP        & $0.9699 \pm 0.0064$ & 0.00 & 10.72  & 0.0000 & 0.5550 \\
\bottomrule
\end{tabular}
\end{table}

\begin{figure}[t]
\centering
\includegraphics[width=0.98\columnwidth,trim=0 0 0 24.9bp,clip]{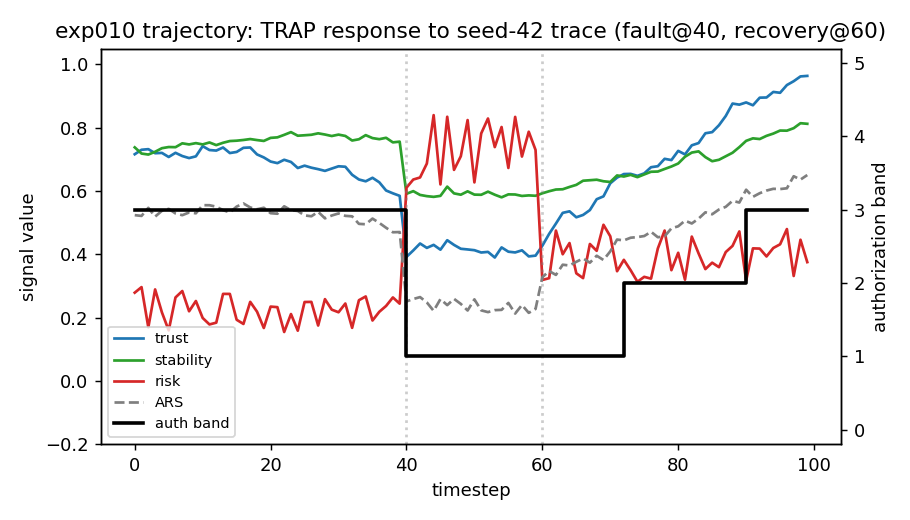}
\caption{TRAP response over the first of the $100$ seed-$42$ traces: trust, stability, risk and
ARS (left axis) with the authorization level (right axis). Dotted lines mark
degradation ($t=40$) and recovery onset ($t=60$).}
\label{fig:trajectory}
\end{figure}

Table~\ref{tab:baseline} reports all five policies. The asymmetry has the intended effect: TRAP restricts in the degradation
interval itself ($\mathrm{RR}=0$; Fig.~\ref{fig:trajectory} shows this on one
trace) while the symmetric rule, whose restriction must also accumulate $k$
observations, takes $2.00$ intervals; and TRAP re-escalates in $10.72$ intervals
against $17.14$, a $37.5\%$ reduction. In this evaluation the faster response does not reduce stability: TRAP
has the highest and least variable AS among adaptive policies. The static
policy's $\mathrm{AS}=1.0000$ reflects only that it never changes level.

The input signal also matters. Risk-only, driven by $1-R_t$, recovers fastest
($2.28$), consistent with $1-R_t$ rising when the risk range falls at recovery
onset, yet grants on average $24.05\%$ of its escalations during degradation,
when $1-R_t$ is lower but redrawn independently each interval. Trust-only recovers slowly ($15.13$). TRAP and the symmetric rule grant no escalation during
degradation ($\mathrm{PER}=0$). TRAP does not dominate on every metric: its
$\mathrm{PRR}$ of $0.5550$ exceeds the symmetric rule's $0.4798$. Because trust drifts upward through the recovery phase, earlier re-escalation
tends to occur at lower trust; the PRR gap is consistent with faster recovery.

Table~\ref{tab:paired} compares TRAP and the symmetric rule trace by trace. All
three differences have intervals that exclude zero. Both policies re-escalate on
$99$ of the $100$ traces; on the remaining one the symmetric rule does not
re-escalate before the horizon ends, while TRAP does after $7$ intervals, so
that trace has no paired RT or PRR. On the $99$, TRAP re-escalates $6.38$
intervals earlier on average: earlier on $34$, equal on $65$ and never later, so
the mean advantage comes from about one third of the traces. TRAP's PRR is
higher on $19$ of the $99$ and never lower. RR has no sampling spread: it is $0$
for TRAP on all $100$ traces and $2$ for the symmetric rule on all $98$ where it
is defined. TRAP's AS also exceeds that of trust-only, with interval
$[+0.0010,\,+0.0042]$, and that of risk-only, with $[+0.0285,\,+0.0345]$. Its PRR,
however, cannot be distinguished from trust-only's ($[-0.023,\,+0.081]$, $95$
paired traces). The public package gives per-policy intervals for every metric;
RT, for example, is $[9.10,\,12.49]$ for TRAP and $[15.38,\,18.93]$ for the
symmetric rule.

\begin{table}[t]
\caption{Paired differences, TRAP minus Symmetric, on the shared traces
(exp010). $n$: traces on which both values are defined; the trace missing from
RT and PRR is one on which only TRAP re-escalates. Counts: traces on which TRAP
is lower / equal / higher.}
\label{tab:paired}
\centering
\footnotesize
\setlength{\tabcolsep}{4pt}
\begin{tabular}{@{}lrrcc@{}}
\toprule
\textbf{Metric} & $n$ & \textbf{Mean} & \textbf{95\% interval} & \textbf{Counts} \\
\midrule
AS  & 100 & $+0.0053$ & $[+0.0033,\,+0.0072]$ & 19 / 21 / 60 \\
RT  &  99 & $-6.38$   & $[-8.28,\,-4.57]$     & 34 / 65 / \phantom{0}0 \\
PRR &  99 & $+0.0707$ & $[+0.0421,\,+0.1027]$ & \phantom{0}0 / 80 / 19 \\
\bottomrule
\end{tabular}
\end{table}

\subsection{Parameter Sensitivity}
\label{sec:res-sensitivity}

\begin{table}[t]
\caption{One-factor sensitivity of TRAP ($n=100$, seed $42$). Baseline row in
each block is $w=(0.5,0.3,0.2)$, $\theta_e=0.05$, $\theta_r=0.10$, $k=3$.
$\mathrm{RR}=0.00$ in every configuration.}
\label{tab:sensitivity}
\centering
\footnotesize
\begin{tabular}{@{}llcc@{}}
\toprule
\textbf{Sweep} & \textbf{Setting} & \textbf{AS} & \textbf{RT} \\
\midrule
\multirow{4}{*}{Weights $w$}
 & $(0.3,0.4,0.3)$ & 0.9730 & \phantom{0}9.99 \\
 & $(0.4,0.4,0.2)$ & 0.9720 & 12.58 \\
 & $(0.5,0.3,0.2)$ & 0.9699 & 10.72 \\
 & $(0.6,0.2,0.2)$ & 0.9699 & 11.35 \\
\midrule
\multirow{3}{*}{$\theta_e$}
 & 0.03 & 0.9686 & \phantom{0}9.18 \\
 & 0.05 & 0.9699 & 10.72 \\
 & 0.07 & 0.9716 & 12.57 \\
\midrule
\multirow{3}{*}{$\theta_r$}
 & 0.05 & 0.9699 & 10.72 \\
 & 0.10 & 0.9699 & 10.72 \\
 & 0.15 & 0.9699 & 10.72 \\
\midrule
\multirow{3}{*}{$k$}
 & 2 & 0.9694 & \phantom{0}9.48 \\
 & 3 & 0.9699 & 10.72 \\
 & 4 & 0.9706 & 11.74 \\
\bottomrule
\end{tabular}
\end{table}

\begin{figure}[t]
\centering
\includegraphics[width=0.98\columnwidth,trim=0 0 0 25.4bp,clip]{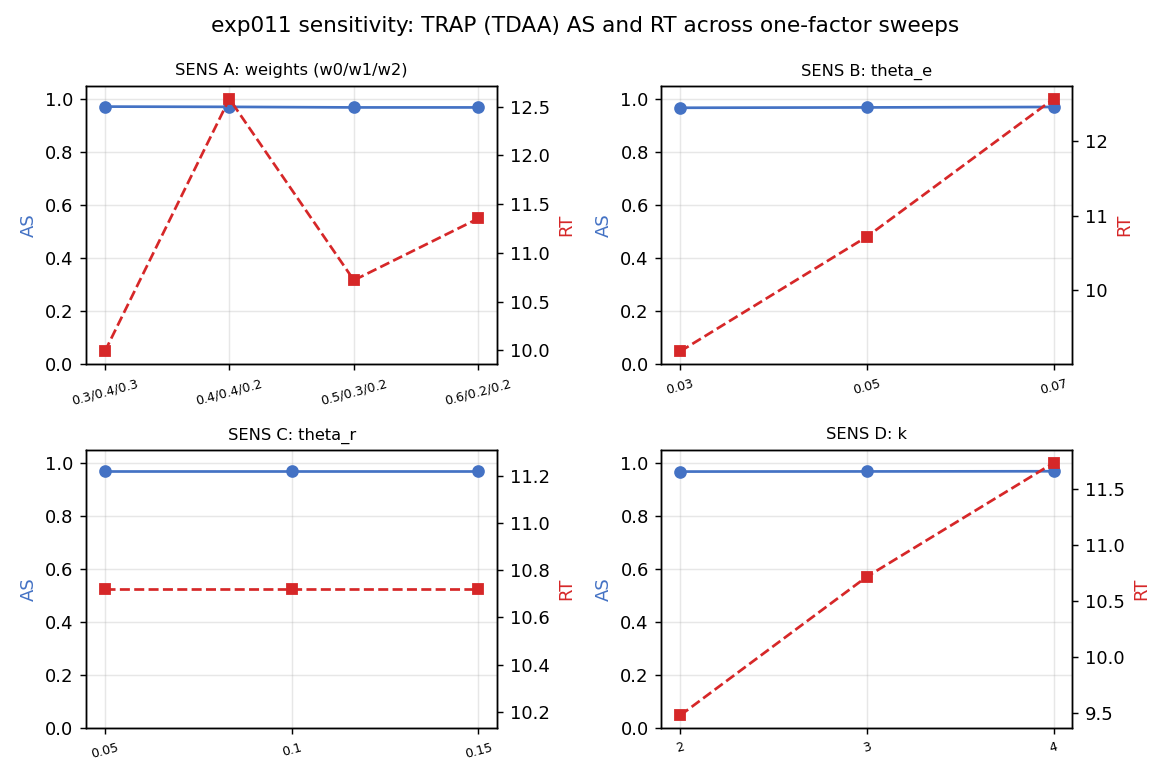}
\caption{AS (left axes) and RT (right axes) across the four one-factor sweeps.}
\label{fig:sensitivity}
\end{figure}

Across all $13$ sweep points (Table~\ref{tab:sensitivity},
Fig.~\ref{fig:sensitivity}), $\mathrm{RR}=0$ throughout and AS varies only within
$[0.9686,\,0.9730]$. The $13$ points comprise $10$ distinct configurations, since
each block repeats the baseline. Within the evaluated grid, RT rises
monotonically with $\theta_e$ ($9.18 \to 12.57$) and with $k$
($9.48 \to 11.74$); both parameters require more evidence, or a longer run of
it, before re-escalation. $\theta_r$, which appears only in the restriction
branch, leaves RT at exactly $10.72$.
Weights move RT non-monotonically ($9.99$--$12.58$). Because $g$ is a weighted
sum, changing $w$ changes the score by different amounts at different signal
values, and therefore changes where a trace crosses the fixed band edges.

Paired on the shared traces ($98$ to $100$ per step, those on which RT is
defined at both settings), every step in $\theta_e$ and $k$ raises RT, and
each interval excludes zero:
$+1.54\ [+0.77,\,+2.24]$ and $+2.38\ [+1.98,\,+2.80]$ for
$\theta_e = 0.03 \to 0.05 \to 0.07$, and
$+1.24\ [+0.95,\,+1.57]$ and $+1.29\ [+1.08,\,+1.55]$ for $k = 2 \to 3 \to 4$.
RT falls on at most $4$ traces in any of these steps. For the weights,
the first two steps move RT in opposite directions, with intervals
$[+0.56,\,+4.60]$ and $[-3.67,\,-0.18]$. The step from $(0.5,0.3,0.2)$ to
$(0.6,0.2,0.2)$, $+0.88\ [-0.28,\,+1.95]$, cannot be distinguished from zero.
The $\theta_r$ sweep leaves RT unchanged on every trace.

\subsection{Gradual Degradation and Excess Authority}
\label{sec:res-gradual}

\begin{table}[t]
\caption{Excess-authority duration in exp012: mean number of the $20$
degradation intervals with $a_t>\sigma(g_t)$ over all $100$ traces; $L=5$
includes the $4$ traces violating the gradual condition (last row). In
parentheses: traces on which the policy lowers its level during degradation.}
\label{tab:gradual}
\centering
\scriptsize
\setlength{\tabcolsep}{1.3pt}
\begin{tabular}{@{}lcccc@{}}
\toprule
\textbf{Policy} & \textbf{Abrupt} & $\boldsymbol{L=5}$ & $\boldsymbol{L=10}$ & $\boldsymbol{L=20}$ \\
\midrule
Static      & 19.85 \phantom{0}(0)\phantom{0} & 19.05 \phantom{0}(0)\phantom{0} & 17.89 \phantom{0}(0)\phantom{0} & 15.41 \phantom{0}(0)\phantom{0} \\
Trust-only  & 17.24 \phantom{0}(99)            & 19.82 \phantom{0}(7)\phantom{0} & 19.96 \phantom{0}(0)\phantom{0} & 19.86 \phantom{0}(0)\phantom{0} \\
Risk-only   & \phantom{0}4.41 (100)            & \phantom{0}7.46 (100)           & 10.49 (100)                     & 15.28 (100) \\
Symmetric   & \phantom{0}9.79 \phantom{0}(98)  & 10.12 \phantom{0}(98)           & 10.28 \phantom{0}(98)           & \phantom{0}9.85 \phantom{0}(97) \\
TRAP        & \phantom{0}0.81 (100)            & 18.92 \phantom{0}(4)\phantom{0} & 18.15 \phantom{0}(0)\phantom{0} & 15.89 \phantom{0}(0)\phantom{0} \\
\midrule
TRAP$-$Sym. & $-8.98$ & $+8.80$ & $+7.87$ & $+6.04$ \\
95\% int.   & $[-10.10,-7.86]$ & $[+7.72,+9.88]$ & $[+7.09,+8.63]$ & $[+5.51,+6.55]$ \\
Violations  & 100 & 4 & 0 & 0 \\
\bottomrule
\end{tabular}
\end{table}

The ramps change the timing of degradation, not its size: the decline of $g$
from $t_F-1$ to the end of the ramp averages $0.308$ (abrupt) and $0.303$,
$0.308$, $0.308$ ($L=5,10,20$). For $L=10$ and $L=20$ every trace is fully
gradual, with largest one-interval decline inside the ramp $0.094$ and $0.075$.
For $L=5$, $4$ traces still contain a decline above $\theta_r$ (at most
$0.112$), as do all $100$ abrupt traces; these are reported, not removed.

Under abrupt degradation TRAP holds excess authority for $0.81$ of the $20$
degradation intervals ($[0.38,\,1.31]$; Table~\ref{tab:gradual}), fewer than
every other policy and fewer than the symmetric rule on all $100$ traces. Under
the ramps it lowers its level only on the $4$ violating traces, as
Algorithm~\ref{alg:trap} predicts, and its EAD over all $100$ traces is
$18.92$, $18.15$ and $15.89$. This decrease with $L$ contradicts our recorded prediction
that EAD would grow with $L$: with a longer ramp, readiness falls below the
held level later. The symmetric rule restricts under every profile, later as
$L$ grows (mean RR $2.00$, $3.26$, $5.27$, $8.85$), and its EAD stays near
$10$. Over all $100$ traces, TRAP's EAD exceeds the symmetric rule's on $88$,
$96$ and $97$ traces for $L=5,10,20$ and is lower only on $2$ $L=5$ traces, both violating the
condition. On the other $96$ $L=5$ traces, which meet it, TRAP never lowers its
level and its EAD is $19.17$ ($[18.75,\,19.46]$) against the symmetric rule's $10.14$, a paired
difference of $+9.03$ ($[+7.96,\,+10.06]$), never negative. For $L\ge10$ the
overall results are fully gradual; there TRAP is never below the static policy
and is above it on the $5$ traces on which it escalated before onset. Its AS
of $0.997$--$0.998$ under the ramps reflects a level that rarely changes, not
robustness; its RT is defined on only $12$ to $16$ traces and is not compared.

Risk-only lowers its level on every trace in every profile because $1-R_t$ is
redrawn each interval and falls by more than $\theta_r$ inside every ramp.
Trust-only and risk-only map their own signals, $T_t$ and $1-R_t$, which lie
above $g_t$; they exceed $\sigma(g_t)$ in $34.2$ and $39.8$ of the $40$
pre-onset intervals, so their EAD mixes this offset with restriction delay, and
restriction is compared among static, symmetric and TRAP. Against their own
signals (a post-hoc diagnostic) their abrupt-case EAD is $1.65$ and $0.90$,
above TRAP's, and risk-only's EAD rises with $L$ under either reference.

\subsection{Multi-Level Conditions in Ordinary Operation}
\label{sec:res-prevalence}

\begin{figure}[t]
\centering
\includegraphics[width=0.92\columnwidth,trim=0 0 0 23.2bp,clip]{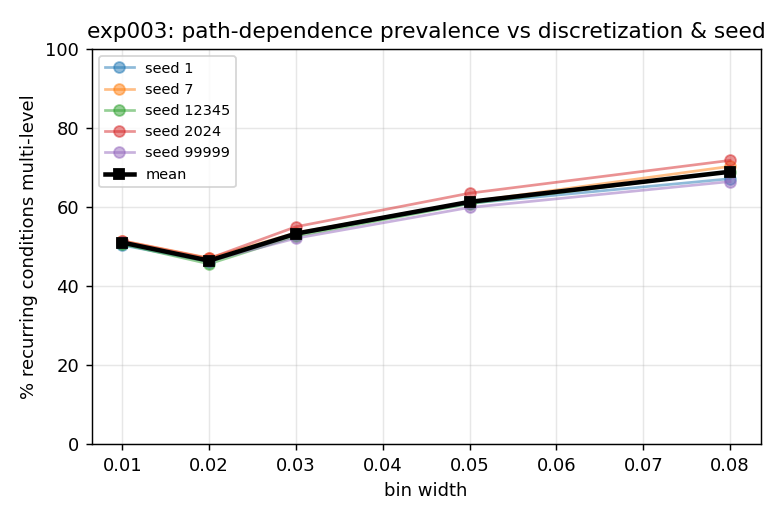}
\caption{Share of recurring conditions observed at two or more authorization
levels against bin width, per seed and averaged.}
\label{fig:prevalence}
\end{figure}

This analysis supplies the condition grid of Section~\ref{sec:res-pdi} and
shows, from ordinary TRAP traces rather than constructed histories, that
identical conditions recur at different authorization levels. Mining $4{,}000$ ordinary traces per setting over five bin widths and five
seeds, the share of recurring conditions observed at two or more levels is
$50.97\%$, $46.40\%$, $53.23\%$, $61.25\%$ and $68.87\%$ for bin widths $0.01$ to
$0.08$ (Fig.~\ref{fig:prevalence}); across seeds it varies by at most $5.35$
points, so the effect is not seed-specific. The increase with bin width is a coarsening effect; the more informative figure
is the $46$--$51\%$ floor at the finest bins. This prevalence partly reflects level persistence within single episodes,
so it is not itself a path-dependence measurement; separating the two requires
the measurement of Section~\ref{sec:res-pdi}.

\subsection{The Structural Result Used Here}
\label{sec:res-pdi}\label{sec:res-reach}\label{sec:res-pilot}\label{sec:res-gen}

Here we restate the structural result established
in~\cite{companion}. Exhaustive enumeration there finds $13$ of $18$
level--counter states reachable under the envelope: level $5$ is excluded,
since entering it needs $\mathrm{up}(4)+\theta_e=0.850>0.770$, while level $4$
is reachable with a zero counter. A pilot generator that approached targets by
ramps returned $\widehat{\mathrm{HSF}}=0.000$ (H1 not supported); enumeration showed carried
level $3$ was reachable but never produced, identifying a coverage failure of the
instrument rather than a property of the policy, and the generator of
Section~\ref{sec:pd} replaced it. That pilot outcome concerns the discarded
instrument only; every value below comes from the verified final generator.

With that generator, carried levels $\{0,\dots,4\}$ were requested at each of the
$50$ targets ($g^\star \in [0.484,0.548]$) under seeds $42$ and $137$. Of $500$
requests, $400$ succeeded; the $100$ failures are exactly the level-$4$ requests.
Their cause is the construction, not the envelope: escalating into level $4$
needs $\mathrm{up}(3)+\theta_e=0.700$, but with $S=0.75$ and $R$ at its minimum
the generator reaches only $0.5(1.0)+0.3(0.75)-0.2(0.150)=0.695$, which is below
the $0.700$ required, although the envelope permits $0.770$. Conditional on
this verified generator, readiness region and construction, the result is
uniform: $\widehat{\mathrm{HSF}}=1.000$ and mean
$\widehat{\mathrm{PDI}}=3.000$, every target at $3$ (H1 supported), identically under both
seeds. The hold branch explains it: at these targets terminal readiness lies
below $\mathrm{up}(a)+\theta_e$ for each carried level and the final step is no
rapid drop, so the carried level passes through unchanged (in $66$ of $100$
target--seed pairs exactly; in $34$, carried level $1$ escalates once to $2$). At
$x^\star=(0.68,0.76,0.16)$, $g^\star=0.536$, agents carrying levels $0$ and $3$
end at levels $0$ and $3$.

Two qualifications apply. The constant $3$ reflects one readiness region, not
the policy's whole domain; and because the envelope admits carried level $4$ (an admissible
history reaches it at the same target), $3$ is a lower bound on the
available spread, not a ceiling.

\section{Discussion}
\label{sec:discussion}

One pattern in the results needs explanation: a single design choice changes
restriction and recovery latency substantially but changes AS by less than one
percentage point. Within the evaluated configurations, Algorithm~\ref{alg:trap}
accounts for this. Both latencies depend on how accumulated state is handled, which the asymmetric
rule does differently in each direction: escalation must rebuild $k$
consecutive observations, restriction discards them outright. AS counts
transitions, not their timing, which is consistent with its small variation. The sensitivity results fit the same branch structure: $\theta_r$, which
parameterizes the discarding branch, leaves recovery unchanged, whereas $\theta_e$
and $k$, which parameterize the rebuilding branch, shift it monotonically; the
weights shift it non-monotonically. PER is not
explained by the transition rule: the three policies applying TRAP's rule range
from $0.0000$ to $0.2405$ depending on their input signal, while both policies
driven by the composite score, asymmetric and symmetric, reach $0.0000$.

exp012 shows the other side of the asymmetry. TRAP's restriction is triggered
by how fast readiness falls, not by how low it is, so a decline that is slow
enough never triggers it, and the hold branch then keeps the level. The
symmetric rule compares readiness with a band edge and therefore restricts under
every profile, at a delay that grows with the ramp. A rule with both triggers,
immediate restriction on a sharp drop and a level-based fallback, is an obvious
candidate for keeping TRAP's abrupt-case behavior while bounding excess
authority; we have not evaluated it.

A separate consequence of the same accumulated state is established
in~\cite{companion}: because carried state persists through the
hold branch, authority depends on history. Removing carried state could avoid
this dependence, but the immediate restriction and conservative recovery measured
here would then have to be preserved by some other mechanism; we do not evaluate
any such alternative. Three
practical implications
follow for deployment; they are hypotheses from a synthetic simulation with
abstract levels, not observations of agent behavior. Under this policy an authorization level is not a reliable
summary of present condition: one condition can be compatible with up to four
levels under the tested construction. Disparities between identically placed
agents need not decay, since the hold branch propagates the carried level while
readiness stays between band edges. Finally, the policy limits history-dependent
disparity only through its finite six-level output range; it imposes no tighter
bound.

Future work may extend the proposed runtime authorization approach to
security-critical application domains. In particular, secure mobility provides
a relevant application domain for evaluating trust-responsive authorization
under dynamic operational conditions~\cite{securemobility}. Furthermore,
healthcare data management offers opportunities to investigate
privacy-preserving access control, particularly through encryption and
biometric authentication mechanisms~\cite{healthrecords}. Cloud computing
represents another relevant domain, where explainable AI can support transparent
task classification and policy evaluation~\cite{cognitiveztna}. These
applications provide opportunities to examine authorization behavior across
different operational contexts. They also enable further investigation of
policy consistency, parameter sensitivity, and runtime adaptation under varying
security and trust requirements.

An extended version of the present conference article is also available in a
related research publication~\cite{verdictprotocol}. Shihab et al. propose a
circularity-controlled evaluation methodology for runtime governance of agentic
AI. This related work provides further methodological context for evaluating
runtime governance mechanisms. It also offers a basis for examining the
relationship between authorization policies, governance decisions, and
empirical evaluation. Consequently, the two studies can be considered
complementary contributions to trustworthy runtime governance of agentic AI.

\section{Limitations}
\label{sec:limits}

\begin{enumerate}
\item \textit{One random stream.} All behavioral results, including exp012, come
from one seed-$42$ stream of $100$ independent traces, shared by every policy,
setting and profile. The
bootstrap intervals characterize trace-level sampling uncertainty under the fixed
generator configuration. They do not establish robustness across independent
random seeds, alternative signal configurations, or noise and degradation
regimes. Event-metric means rest on the traces in which each event
occurred, which differ slightly between policies.
\item \textit{Degradation scenarios.} TRAP's restriction advantage holds for
abrupt degradation only (Section~\ref{sec:res-gradual}). exp012 tests linear
ramps of one severity range at three lengths; stepwise, oscillating, partial or
repeated degradation is untested. RR, RT, PER and PRR are defined against known
onsets $t_F$ and $t_R$ and do not describe unannounced or repeated degradation.
\item \textit{Recovery not gated on trust.} The mean share of TRAP's
post-recovery escalations occurring while $T_t<0.60$ is $0.555$, compared with
$0.480$ for the symmetric rule; the paired difference is $+0.0707$ (95\%
bootstrap interval
$[0.0421,\,0.1027]$; Tables~\ref{tab:baseline}--\ref{tab:paired}). The threshold
$0.60$ is an evaluation choice, and a trust-based recovery condition is not
evaluated.
\item \textit{Metric resolution.} EAD is measured against $\sigma(g_t)$ with the
baseline weights and fixed band edges, only within $[t_F,t_R)$, and only in
exp012; it counts intervals, not the permissions the excess level would grant.
Its reference is the readiness mapping the $g$-driven policies themselves use.
EAD therefore measures readiness-relative excess authority, not an independent
real-world safety outcome. This is related to the evaluation-circularity concern
discussed in~\cite{cce}.
$\mathrm{RR}=0$ in every TRAP configuration, so the design cannot discriminate
settings on restriction latency; the $\theta_r$ invariance concerns recovery
only.
\item \textit{Policy and sensitivity scope.} One asymmetric rule is studied,
not asymmetric authorization in general; sweeps vary one factor over three or
four values, so interactions are unmeasured and the monotonicity in $\theta_e$
and $k$ holds over the evaluated points only. The band edges are not varied.
\item \textit{Synthetic signals, no agent.} Trust, stability and risk are
synthetic, TRAP has not been run with a deployed or LLM-based agent, and its
levels carry no concrete permissions here. No real-world validation is claimed;
behavior may differ under empirical signal dynamics.
\item \textit{Scope of the borrowed structural result.} It concerns one rule at
fixed parameters, one readiness region, one construction mode and $50$
conditions, and is a lower bound; the prespecified H2 is untested.
\end{enumerate}

\section{Reproducibility and Code Availability}
\label{sec:repro}

The code, stored results and a pinned environment are publicly available under
the MIT License at \url{https://github.com/ahsan-h-shihab/trap}. The version
described here is commit
\href{https://github.com/ahsan-h-shihab/trap/tree/02037324115480e5ac4d89a20000ae05f53dd20c}{\code{020373241154}};
later versions may differ. In it, the \code{frozen} directory contains the
frozen artifact. This version is archived on Zenodo as TRAP software release
1.0.0, doi:~\href{https://doi.org/10.5281/zenodo.23836060}{\nolinkurl{10.5281/zenodo.23836060}}~\cite{trapzenodo}.
Separate immutable \mbox{SHA-256} manifests cover the core policy,
envelope and path-dependence experiments and the behavioral experiments, and
TRAP keeps its legacy code identifier, \code{tdaa}, for harness compatibility. Four verifiers check the frozen experiment directories against
their recorded digests, and a reproduction harness passes all $31$ checks
covering the envelope, reachability, condition grid, path-dependence statistics
and artifact integrity. We independently re-derived the behavioral and
sensitivity results, including all $500$ stored per-run records, and confirmed
byte-for-byte agreement of the figure data. The interval estimates of
Section~\ref{sec:results} were computed after the freeze by a script that,
before computing any interval, verifies that the per-run records reproduce every
stored aggregate exactly. The exp012 checker confirms that its generator with
$L=1$ reproduces the frozen generator bit for bit and that the abrupt control
reproduces all $500$ stored exp010 per-run records, then regenerates every
stored exp012 result and compares it exactly. Both analyses use a fixed
bootstrap seed and are deterministic, and neither modifies the frozen tree. The
script \code{run\_all\_checks.py} runs all of these checks; by default it
verifies the results of Section~\ref{sec:res-prevalence} against their digests,
and with \code{--exp003} it also regenerates them.

\section{Conclusion}

We have empirically characterized the behavior of a trust-responsive runtime
authorization policy. On the tested traces with abrupt degradation, TRAP
restricts authority in the interval in which degradation occurs and first
re-escalates $37.5\%$ sooner than a symmetric rule, and in this evaluation it
has the highest and least variable authorization stability among the adaptive
policies, while, within the evaluated configurations, recovery time increases
monotonically with $\theta_e$ and $k$. Within the simulator, paired intervals
over the shared traces support these latency and mean-stability differences. The
cost is visible in two places: a higher share of post-recovery escalations
occurring below the selected trust threshold and, when the same degradation
arrives gradually, a restriction branch that does not fire, so that TRAP holds excess authority for most of the
degradation phase, longer than the symmetric rule. The evidence therefore
supports rapid restriction under abrupt degradation, not reliable protection
under gradual degradation. The principal limitations are a synthetic testbed
with one random stream, one generator configuration and no agent; authorization
levels without concrete permissions; and a small set of degradation profiles.
A further question is whether a rule can keep immediate restriction
while bounding both excess authority under gradual decline and
history-dependent disparity, for example through a level-based fallback trigger
or by decaying carried level under sustained stable conditions; the metrics used
here, including EAD, would show whether such a change preserves the behavior
reported above.

\section*{Acknowledgment}

The authors acknowledge support from project 2020-1.1.2-PIACI-KFI-2020-00129
and BR24993001, ``Creating a Large Language Model (LLM) to support the Kazakh
language and technological progress,'' provided by the Committee of Science of
the Ministry of Science and Higher Education of the Republic of Kazakhstan.
This work was also supported by the National Key Research and Development
Program of China (2013YFB3406801). Authors also aknowledge that Anthropic Claude was used as an assistive tool for software implementation, code
debugging, \LaTeX{} preparation, and related technical tasks. AI also used for minor langauge editing. The
authors reviewed and verified all outputs and take full responsibility for the
scientific content, analyses, interpretations, and results.


\end{document}